\documentclass{article}

\usepackage{arxiv}

\usepackage[utf8]{inputenc} % allow utf-8 input
\usepackage[T1]{fontenc}    % use 8-bit T1 fonts
\usepackage{hyperref}       % hyperlinks
\usepackage{url}            % simple URL typesetting
\usepackage{booktabs}       % professional-quality tables
\usepackage{amsmath}
\usepackage{amsfonts}       % blackboard math symbols
\usepackage{nicefrac}       % compact symbols for 1/2, etc.
\usepackage{microtype}      % microtypography
\usepackage{cleveref}       % smart cross-referencing
\usepackage{lipsum}         % Can be removed after putting your text content
\usepackage{graphicx}
\usepackage[numbers]{natbib}
\usepackage{doi}
\usepackage{makecell}

\title{Large Language Models for Code Summarization}
\date{}
\author{
        \href{https://orcid.org/0000-0002-0458-2576}{\includegraphics[scale=0.06]{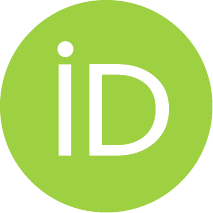}\hspace{1mm}Balázs Szalontai}\\
	\texttt{bukp00@inf.elte.hu} \\
	\And
        \href{https://orcid.org/0009-0008-9782-3869}{\includegraphics[scale=0.06]{orcid.pdf}\hspace{1mm}Gergő Szalay}\\
	\texttt{d5ij3p@inf.elte.hu} \\
        \And
        \href{https://orcid.org/0009−0000−1936−0207}{\includegraphics[scale=0.06]{orcid.pdf}\hspace{1mm}Tamás Márton}\\
	\texttt{nru0i6@inf.elte.hu} \\
        \And
        \href{https://orcid.org/0009-0002-5239-0517}{\includegraphics[scale=0.06] {orcid.pdf}\hspace{1mm}Anna Sike}\\
	\texttt{ci6u9i@inf.elte.hu} \\
        \And
        \href{https://orcid.org/0000-0003-3431-0667}{\includegraphics[scale=0.06]{orcid.pdf}\hspace{1mm}Balázs Pintér}\\
	\texttt{pinter@inf.elte.hu} \\
        \And
        \href{https://orcid.org/0000-0002-9503-9623}{\includegraphics[scale=0.06]{orcid.pdf}\hspace{1mm}Tibor Gregorics}\\
	\texttt{gt@inf.elte.hu}
}

\renewcommand{\headeright}{Technical Report}
\renewcommand{\undertitle}{Technical Report}
\renewcommand{\shorttitle}{Large Language Models for Code Summarization}

\hypersetup{
pdftitle={Large Language Models for Code Summarization},
pdfsubject={Large Language Models, Code generation, Code explanation},
pdfauthor={Balázs Szalontai, Gergő Szalay, Tamás Márton, Anna Sike, Balázs Pintér, Tibor Gregorics},
pdfkeywords={Large Language Models, Code explanation, Code generation},
}

\begin{document}
\maketitle

\vspace{-1cm}
\begin{center}
	{
	\large
	Eötvös Loránd University\\Faculty of Informatics
	}
\end{center}
\vspace{1cm}

\begin{abstract}
  Recently, there has been increasing activity in using deep learning for
  software engineering, including tasks like code generation and summarization.
  In particular, the most recent coding Large Language Models seem to perform
  well on these problems. In this technical report, we aim to review how these
  models perform in code explanation/summarization, while also investigating
  their code generation capabilities (based on natural language descriptions).
\end{abstract}

% keywords can be removed

\newpage
\tableofcontents

\section{Introduction}
\label{section:intro}

The introduction of Encoder-Decoder architectures in natural language processing
\cite{sutskever2014sequence} (both recurrent \cite{cho2014learning} and
Transformer-based \cite{vaswani2017attention}) has motivated researchers to
apply them to software engineering. One important application is generating
summaries of code \cite{shido2019automatic, aljumah2022bi,
  iyer2016summarizing}. A code summarization tool is useful for example to
understand legacy code or to create documentation. Since the spread of Large
Language Models (LLMs), the working programmer has many more opportunities to
use deep learning-based tools. Closed models (such as GPT-4
\cite{openai2024gpt4} or Gemini \cite{geminiteam2023gemini}) and open models
(such as CodeLlama \cite{rozière2023code} or WizardCoder
\cite{luo2023wizardcoder}) demonstrate impressive capabilities of generating
source code based on a task description, as well as generating natural-language
summary of code.

The main objective of this technical report is to investigate how well
open-sourced LLMs handle source code in relation with natural language text. In
particular, we discuss results of some of the most acknowledged open-source
LLMs, focusing on their code summarization/explanation (code-to-text)
capabilities. We also discuss code generation (text-to-code) capabilities of
these LLMs, as this is often considered to be their most defining capability.
That is, LLMs are often ranked simply based on results on a code generation
benchmark. Benchmarking datasets for measuring code generation capabilities
include HumanEval \cite{chen2021evaluating}, APPS
\cite{hendrycks2021measuring}, MBPP \cite{austin2021program} and DS-1000
\cite{lai2022ds1000}. For measuring code summarization or explanation
capabilities, fewer benchmarks have been published, such as CodeXGLUE
\cite{lu2021codexglue} and HumanEvalExplain \cite{muennighoff2023octopack}.

\Cref{benchmarking} describes evaluation metrics and benchmark 
datasets, used for measuring code generation and explanation performance of LLMs. \Cref{open}
reviews some of the prominent open-source LLMs, discussing their capabilities of synthesizing 
and explaining code. Finally, \Cref{conclusion} concludes our report.

\section{Benchmarking LLMs}
\label{benchmarking}

We review results of various LLMs on some widely acknowledged benchmarks. In
this report, we focus on two benchmark tasks: (i) code generation and (ii) code
summarization/explanation. Before reviewing these tasks and their benchmarks, we
describe the metrics used for evaluation.

\subsection{Metrics}

Before describing the various benchmark datasets, we outline the metrics that are 
used for measuring performance on these datasets.

\subsubsection{Pass@k}

In the context of LLMs, perhaps the most noteworthy metric is the pass@k
performance. It was introduced by Kulal et al. \cite{kulal2019spoc}. The LLM is
prompted to solve some kind of a task. The integer $k$ denotes the number of
generated responses (i.e. attempts) per prompt. The execution of the task in the
prompt is considered successful if there is at least one correct response among
the generated responses (which is usually validated using unit tests). In
theory, the total fraction of problems solved should be reported as the result
of this benchmark. In practice however, in order to decrease the variance of the
results, a good trick is to prompt the model $n$ ($\geq k$) times, and let $c$ be
the number of correct responses. This way, the pass@k performance can be
estimated as
$$pass@k := \prod_{problems}[1-\frac{\binom{n-c}{k}}{\binom{n}{k}}]$$

\subsubsection{BLEU}

The BLEU score \cite{papineni2002bleu} was originally designed for evaluating
translation techniques (including Neural Machine Translation). It attempts to
capture a numerical metric of how close a generated text is to the goal. The
ground truth is generally a set of good solutions (since for example for
translation, there are almost always multiple ways to perfectly translate the
same sentence). To calculate the BLEU score, n-grams (sequences of n words) of
the generated sequence have to be compared with the set of goals. The BLEU score
measures how many of the generated n-grams match those in the goals, considering
precision and brevity penalty. The higher the BLEU score, the better the
generated text is considered to be. There have been multiple proposed variants
of the BLEU score, one of which is the smoothed BLEU \cite{lin2004orange}.

\subsubsection{ROUGE}

ROUGE \cite{lin2004rouge} is a set of measures to automatically determine the
quality of a generated text, and functions similarly to BLEU. It compares the
generated text to ideal texts created by humans. It offers more options for
comparing with multiple ground truths: ROUGE-N (an n-gram recall), ROUGE-L
(longest common subsequence), ROUGE-W (weighted longest common subsequence),
ROUGE-S and SU (Skip-Bigram Co-Occurrence Statistics).

\subsection{Code generation (text-to-code)}

The most frequently highlighted aspect of coding LLMs is their code generation
capability. Results on datasets such as HumanEval are usually used for ranking 
different models. These results are visualized and kept up-to-date on leaderboards, 
which allow for obtaining recent information about current LLMs and their 
capabilities to synthesize code. Two of such leaderboards are the \textbf{Big Code 
Models Leaderboard}\footnote{https://huggingface.co/spaces/bigcode/bigcode-models-leaderboard}
and the \textbf{CanAiCode Leaderboard}\footnote{https://huggingface.co/spaces/mike-ravkine/can-ai-code-results}. 

\subsubsection{HumanEval and its variants}

HumanEval \cite{chen2021evaluating} is a benchmark that contains 164
handwritten programming problems. Each problem includes a Python function
signature, docstring, body, and on average 7.7 unit tests per problem. The goal
of the model is to synthesize a functionally correct function body.

Multiple variants of the HumanEval benchmark have been proposed. HumanEval+
\cite{liu2023code} extends the number of test cases by 80x. The additional test
cases revealed that many models were initially slightly misjudged (and often
overpraised) when running just the original test cases for each problem. Another
extension is HumanEvalSynthesize \cite{muennighoff2023octopack}, which extends
the HumanEval benchmark to multiple programming languages (JavaScript, Java, Go,
C++, Rust). Another extension is HumanEval-XL \cite{peng2024humanevalxl}, which
extends also the number of natural languages (to 23), and programming languages
to 12. This extension provides 22080 prompts in total with 8.33 test cases for
validation on average.

\subsubsection{APPS}

The Automated Programming Progress Standard (APPS) benchmark
\cite{hendrycks2021measuring} contains $10{,}000$ problems: simple introductory
problems, interview-level problems, and coding competition challenges. The data
is separated evenly into training and test sets, with 5000 problems each.
Evaluating code generation capabilities of models (fine-tuned on the training
set) is facilitated by a large bank of test cases: 21.2 on average per problem.
The programs are gathered from openly accesible sites, such as Codewars,
AtCoder, Kattis, and Codeforces.

\subsubsection{MBPP}

MBPP \cite{austin2021program} is a benchmark designed to measure the ability to
synthesize short Python programs from natural language descriptions. It contains
974 programming tasks, designed to be solvable by entry-level programmers. The
problem solutions can be mathematical in nature (58\%), or involve list
processing (43\%), string processing (19\%), integer sequences (9\%) or the use
of some other data structures (2\%). Test cases are used to check functional
correctness of generated programs (three for each problem).

\subsubsection{DS-1000}

DS-1000 \cite{lai2022ds1000} is a code generation benchmark with a thousand data 
science problems spanning seven Python libraries: NumPy, Pandas, TensorFlow, PyTorch, 
SciPy, Scikit-learn, and Matplotlib. On average, a problem is evaluated using 1.6 
test cases. The problems included in the dataset originate from 451 unique StackOverflow 
problems. To defend against potential memorization, more than half of the DS-1000 
problems are modified from the original StackOverflow problems.

\subsection{Code summarization/explanation (code-to-text)}

\subsubsection{CodeXGLUE}

CodeXGLUE \cite{lu2021codexglue} is a benchmark dataset for program
understanding and generation that includes 14 datasets across 10 tasks. It can
be used for benchmarking performance in a wide range of tasks, such as clone
detection, code completion, natural language code search, or code summarization.
Here, we focus on code summarization.

For code summarization, the CodeSearchNet dataset \cite{husain2020codesearchnet} 
is used. This dataset contains programs in multiple languages: Python, Java, PHP, 
JavaScript, Ruby and Go. For evaluating the summaries, the smoothed BLEU score 
is used.

\subsubsection{HumanEvalExplain}

The HumanEvalExplain benchmark is part of the HumanEvalPack
\cite{muennighoff2023octopack} and aims to determine code explanation
capabilities of large language models. Instead of measuring BLEU or ROUGE
scores, it uses a different strategy. First, given a correct code function, the
model is prompted to generate an explanation of the code. Subsequently, the same
model is tasked to generate the code from scratch given only its own
explanation. The outcome of the second step is measurable by the pass@k metric.
The result on this metric is considered to be the explanation capability of the
model.

\section{Open-sourced LLMs for code}
\label{open}

\begin{figure}[b]
    \centering
    \includegraphics[width=0.98\textwidth]{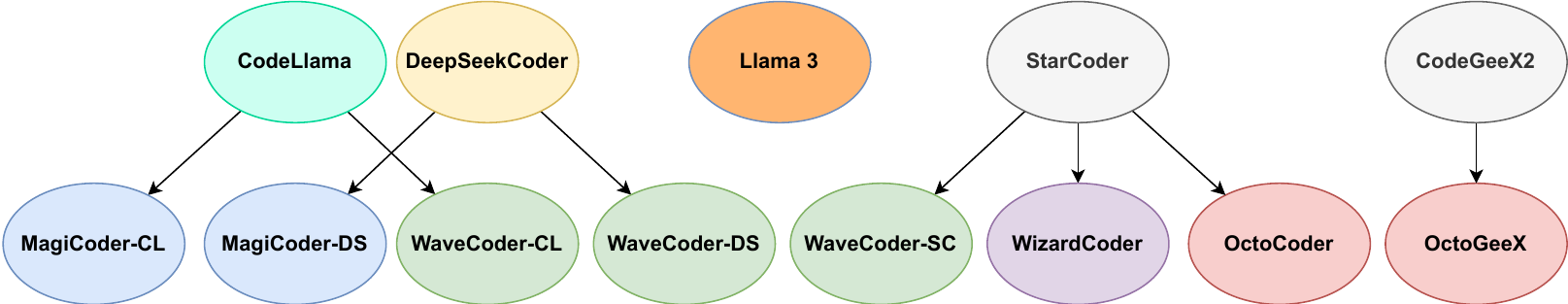}
    \caption{The LLMs we review in this report. If a model was obtained by
    fine-tuning, it is connected to its base model. Families of models are
    highlighted using the same color, while StarCoder and CodeGeeX2 are gray 
    indicating that they are not discussed in this report.}
    \label{fig:llms}
\end{figure}

Now we turn to specific LLMs for code. In general, these models are designed to
solve software engineering-related problems, such as code generation, code
completion, or explaining code. We highlight reported benchmark results on code
generation and explanation. For the benchmarks that report results on multiple
languages, we review them on Python alongside the average results across all
languages. The discussed models and the connections between them are visualized
in \Cref{fig:llms}.

We also discuss the two very recently published Llama3 models (8B and 70B).
Although they should be considered general-purpose LLMs, they were also trained
on code and show very promising capabilities on code generation.

\subsection{OctoCoder and OctoGeeX}

% Summary
OctoCoder and OctoGeeX constitute the LLMs in the OctoPack
\cite{muennighoff2023octopack}. One goal of the authors is to offer
instruction-tuned variants of existing base models. While fine-tuning, they
heavily use Git commit data, which they also publish as the CommitPack. They
fine-tune two base models: StarCoder-16B \cite{li2023starcoder} (obtaining
OctoCoder) and CodeGeeX2-6B \cite{zheng2023codegeex} (obtaining OctoGeeX). The
authors release the HumanEvalPack, which expands on the HumanEval benchmark to a
total of three coding tasks: code repair, code explanation, code generation.
Upon publication, their models achieved the best performance on each benchmark
of the HumanEvalPack among all permissive models: OctoCoder achieved \textbf{46.2\%} on
HumanEvalSynthesize and \textbf{35.1\%} on HumanEvalExplain.

% HumanEval(Synthesize)
In the HumanEvalPack, HumanEvalSynthesize is the benchmark that resembles the
original HumanEval. OctoGeeX achieves \textbf{44.7\%} zeroshot pass@1 performance on
Python and \textbf{30.9\%} across multiple languages. OctoCoder achieves \textbf{46.2\%} zeroshot
pass@1 performance on Python and \textbf{35.5\%} across multiple languages.

% HumanEvalExplain
Among the released benchmarks, HumanEvalExplain is the most relevant to this
report as it measures code explanation capabilities. On this benchmark, OctoGeeX
achieves result of \textbf{30.4\%}, while OctoCoder achieves \textbf{35.1\%}.
% Competing open models at the time of publishing achieve results up to 25.4\%.

\subsection{CodeLlama}

% Summary

CodeLlama \cite{rozière2023code} is the openly accessable Llama2
\cite{touvron2023llama} fine-tuned for programming-related tasks. Alongside the
base model, CodeLlama-Python and CodeLlama-Instruct were also released. These
models come in different sizes: 7B, 13B, 34B and 70B. Some of the models were
also trained for the objective of infilling, which can be used for example for
docstring generation, which is very relevant to the topic of code summarization.
CodeLlama achieved state-of-the-art performance among open models upon
publication on HumanEval (\textbf{67\%}) and MBPP (\textbf{65\%}). It also performs considerably
well on the CodeXGLUE benchmark (\textbf{21.15 BLEU}), which indicates good program
summarization capabilities.

% HumanEval, MBPP, APPS
CodeLlama's code generation capabilities are reported on HumanEval and MBPP.
Notably, the instruction-tuned variant with 70B parameters achieves \textbf{67.8\%}
pass@1, \textbf{90.3\%} pass@10 and \textbf{97.3\%} pass@100 performance on HumanEval, and the
Python specialist model with 70B parameter achieves \textbf{65.6\%} pass@1, \textbf{81.5\%}
pass@10 and \textbf{91.9\%} pass@100 performance on MBPP. The detailed results can be
observed in \Cref{tab:CodeLlamaHumanEvalMBPP}. The authors also benchmark their
models on APPS. The results of CodeLlama-Instruct models on this benchmark are
shown in \Cref{tab:CodeLlamaAPPS}.

\begin{table}[t]
    \begin{center}
        \begin{tabular}{ l | c | c }
            \makecell[c]{Model} & HumanEval (pass@1, 10, 100) & MBPP (pass@1, 10, 100) \\ \hline \hline
            CodeLlama-7B & 33.5\%, 59.6\%, 85.9\% & 41.4\%, 66.7\%, 82.5\% \\
            CodeLlama-13B & 36.0\%, 69.4\%, 89.8\% & 47.0\%, 71.7\%, 87.1\% \\
            CodeLlama-34B & 48.8\%, 76.8\%, 93.0\% & 55.0\%, 76.2\%, 86.6\% \\
            CodeLlama-70B & 53.0\%, 84.6\%, 96.2\% & 62.4\%, 81.1\%, 91.9\% \\ \hline \hline
            CodeLlama-Instruct-7B & 34.8\%, 64.3\%, 88.1\% & 44.4\%, 65.4\%, 76.8\% \\
            CodeLlama-Instruct-13B & 42.7\%, 71.6\%, 91.6\% & 49.4\%, 71.2\%, 84.1\% \\
            CodeLlama-Instruct-34B & 41.5\%, 77.2\%, 93.5\% & 57.0\%, 74.6\%, 85.4\% \\
            CodeLlama-Instruct-70B & 67.8\%, 90.3\%, 97.3\% & 62.2\%, 79.6\%, 89.2\% \\ \hline \hline
            CodeLlama-Python-7B & 38.4\%, 70.3\%, 90.6\% & 47.6\%, 70.3\%, 84.8\% \\
            CodeLlama-Python-13B & 43.3\%, 77.4\%, 94.1\% & 49.0\%, 74.0\%, 87.6\% \\
            CodeLlama-Python-34B & 53.7\%, 82.8\%, 94.7\% & 56.2\%, 76.4\%, 88.2\% \\
            CodeLlama-Python-70B & 57.3\%, 89.3\%, 98.4\% & 65.6\%, 81.5\%, 91.9\% \\
        \end{tabular}
    \end{center}
    \caption{CodeLlama model variants and their performance on the HumanEval and MBPP benchmarks}
    \label{tab:CodeLlamaHumanEvalMBPP}
\end{table}

\begin{table}[t]
    \begin{center}
        \begin{tabular}{ l | c | c | c }
            \makecell[c]{Model} & \makecell{Introductory \\ (pass@5, 10, 100)} & \makecell{Interview \\ (pass@5, 10, 100)} & \makecell{Competition \\ (pass@5, 10, 100)} \\ \hline \hline
            CodeLlama-Instruct-7B & 24.9\%, 29.4\%, 41.3\% & 6.3\%, 8.4\%, 16.1\% & 1.9\%, 3.0\%, 9.2\% \\
            CodeLlama-Instruct-13B & 24.8\%, 29.8\%, 43.5\% & 7.0\%, 9.2\%, 17.3\% & 1.7\%, 2.5\%, 6.3\% \\
            CodeLlama-Instruct-34B & 19.8\%, 25.9\%, 43.5\% & 5.7\%, 8.0\%, 16.9\% & 1.5\%, 2.3\%, 6.4\% \\
        \end{tabular}
    \end{center}
    \caption{CodeLlama-Instruct variants and their performance on the APPS benchmarks}
    \label{tab:CodeLlamaAPPS}
\end{table}

% CodeXGlue
CodeLlama has been benchmarked on the CodeXGLUE dataset, which measures
performance of code summarization. The 7B model achieved \textbf{20.39-20.37} while the
13B model achieved \textbf{21.05-21.15} BLEU score on this benchmark. The authors
compared this result to the results of InCoder \cite{fried2023incoder},
SantaCoder \cite{allal2023santacoder} and StarCoder \cite{li2023starcoder},
that achieve scores between 18.27 and 21.99. This makes CodeLlama-13B on par
with other state-of-the-art models in code summarization.

% HumanEvalExplain
Yu et al. report \cite{yu2024wavecoder} further results on the code explaining
capabilities of the CodeLlama models, on the HumanEvalExplain benchmark.
According to their report, CodeLlama-Instruct-7B scores \textbf{33.5\%} on Python and \textbf{27.3\%}
across multiple languages, while the 13B variant achieves \textbf{40.2\%} on Python and \textbf{28.2\%}
across multiple languages.

\subsection{WizardCoder}

% Summary
WizardCoder LLMs \cite{luo2023wizardcoder} aim to improve the performance of
instruction-following. The authors utilize the Evol-Instruct method (introduced
by WizardLM \cite{xu2023wizardlm}), which involved evolving existing
instruction data to generate more complex and diverse datasets.
WizardCoder-Python was also released for Python-specific problems. The models
come in different sizes: 7B, 15B, 34B. They use StarCoder
\cite{li2023starcoder} as the base model. The models were evaluated on multiple
code generation benchmarks (HumanEval, MBPP and DS-1000), surpassing all other
open-source Code LLMs upon publication. One of the models was also externally
evaluated on the HumanEvalExplain dataset, reaching \textbf{32.5\%} zeroshot pass@1
performance.

% Code generation
WizardCoder has been benchmarked on multiple code generation benchmark datasets:
HumanEval, MBPP and DS-1000. The results can be seen in
\Cref{tab:WizardCoderBenchmarks}.

\begin{table}[t]
    \begin{center}
        \begin{tabular}{ l | c }
            \makecell[c]{Benchmark} & Result (pass@1) \\ \hline \hline
            HumanEval & 57.3\% \\ \hline
            MBPP & 51.8\% \\ \hline
            \makecell[l]{DS-1000 \\ (Format: completion)} & 29.2\% \\ \hline
            \makecell[l]{DS-1000 \\ (Format: insertion)} & 32.8\% \\
        \end{tabular}
    \end{center}
    \caption{WizardCoder-15B and its performance on three code generation benchmarks}
    \label{tab:WizardCoderBenchmarks}
\end{table}

% HumanEvalExplain
The authors of OctoCoder have evaluated WizardCoder on HumanEvalExplain. Here,
WizardCoder achieves \textbf{32.5\%} (zeroshot, pass@1) performance on Python and \textbf{27.5\%}
across multiple languages. The authors state that the 16B variant was evaluated,
but such a model was never published. Thus they probably are slightly mistaken
and actually refer to the 15B variant.

\subsection{DeepSeekCoder}

% Summary
DeepSeekCoder \cite{guo2024deepseekcoder} is a collection of LLMs, trained from
scratch for software engineering-related problems. It has also been trained for
the task of infilling, which could also enable docstring generation.
Instruction-tuned variants were published alongside the base models. These
models come in different sizes: 1.3B, 6.7B, 7B, 33B. On HumanEval, the models
achieve up to \textbf{79.3\%} on Python and \textbf{69.2\%} across multiple languages. They were
also evaluated on the MBPP and DS-1000 benchmark, reaching \textbf{70\%} and \textbf{40.2\%}
respectively. Detailed results can be seen in
\Cref{tab:DeepSeekCoderHumaEvalMBPPDS1000}.

\begin{table}[t]
    \begin{center}
        \begin{tabular}{ l | c | c | c | c }
            \makecell[c]{Model} & \makecell{HumanEval \\ (Python)} & \makecell{HumanEval \\ (Average)} & MBPP & DS-1000  \\ \hline \hline
            DeepSeekCoder-Base-1.3B & 34.8\% & 28.3\% & 46.2\% & 16.2\% \\ \hline
            DeepSeekCoder-Base-6.7B & 49.4\% & 44.7\% & 60.6\% & 30.5\% \\ \hline
            DeepSeekCoder-Base-33B & 56.1\% & 50.3\% & 66.0\% & 40.2\% \\ \hline
            DeepSeekCoder-Instruct-1.3B & 65.2\% & 48.4\% & 49.4\% & - \\ \hline
            DeepSeekCoder-Instruct-6.7B & 78.6\% & 66.1\% & 65.4\% & - \\ \hline
            DeepSeekCoder-Instruct-33B & 79.3\% & 69.2\% & 70.0\% & - \\ 
        \end{tabular}
    \end{center}
    \caption{DeepSeekCoder variants and their performance on the HumanEval, MBPP and DS-1000 benchmarks}
    \label{tab:DeepSeekCoderHumaEvalMBPPDS1000}
\end{table}

% FIM
Similarly to CodeLlama, DeepSeekCoder was also trained for infilling. The
infilling capability of the models were measured using the Single-Line Infilling
benchmarks \cite{allal2023santacoder}. SantaCoder-1.1B, StarCoder-16B,
CodeLlama-Base-7B and CodeLlama-Base-13B score 44\%-68.3\% on Python and
69\%-75.5\% across multiple languages on this benchmark. The results of
DeepSeekCoder can be seen on \Cref{tab:DeepSeekCoderFIM}.

\begin{table}[b]
    \begin{center}
        \begin{tabular}{ l | c | c }
            \makecell[c]{Model} & Python & Average \\ \hline \hline
            DeepSeekCoder-Base-1.3B & 57.4\% & 70.4\% \\ \hline
            DeepSeekCoder-Base-6.7B & 66.6\% & 80.7\% \\ \hline
            DeepSeekCoder-Base-33B & 65.4\% & 81.2\% \\
        \end{tabular}
    \end{center}
    \caption{DeepSeekCoder variants and their fill-in-the-middle performance, measured on the Single-Line Infilling benchmarks}
    \label{tab:DeepSeekCoderFIM}
\end{table}

% HumanEvalExplain
Although the authors did not report any benchmark results on source code
explanation, the authors of WaveCoder \cite{yu2024wavecoder} did evaluate
DeepSeekCoder-6.7B on such a benchmark: they report the results on
HumanEvalExplain. According to them, DeepSeekCoder achieves \textbf{43.9\%} pass@1
performance on Python and \textbf{34.6\%} across multiple languages.

\subsection{MagiCoder}

MagiCoder \cite{wei2023magicoder} LLMs are further fine-tuned variants of
CodeLlama-7B and DeepSeekCoder-6.7B for instruction-following. The approach of
fine-tuning for instruction-following utilizes OSS-INSTRUCT, which leverages a
powerful LLM to automatically generate new coding problems by drawing
inspiration from open-source code snippets. The MagiCoder models were evaluated
on multiple benchmarks, including DS-1000 (achieving up to \textbf{37.5\%}), HumanEval
(achieving up to \textbf{76.8\%}) and MBPP (achieving up to \textbf{75.7\%}).

The code generation performance of MagiCoder models were evaluated on the HumanEval, 
HumanEval+, MBPP, MBPP+, DS-1000 and MultiPL-E benchmarks. The results on the first 
five benchmarks are summarized in \Cref{tab:MagiCoderCodeGen}.

\begin{table}[t]
    \begin{center}
        \begin{tabular}{ l | c | c | c | c | c }
            \makecell[c]{Model} & HumanEval & HumanEval+ & MBPP & MBPP+ & DS-1000 \\ \hline \hline
            MagiCoder-CL-7B & - & - & - & - & 29.9\% \\ \hline
            MagiCoderS-CL-7B & - & - & - & - & 37.5\% \\ \hline
            MagiCoder-DS-7B & 66.5\% & 60.4\% & 75.4\% & 61.9\% & - \\ \hline
            MagiCoderS-DS-7B & 76.8\% & 70.7\% & 75.7\% & 64.4\% & - \\
        \end{tabular}
    \end{center}
    \caption{MagiCoder variants and their pass@1 code generation performance}
    \label{tab:MagiCoderCodeGen}
\end{table}

% HumanEvalExplain
Although the authors did not report any benchmark result on source code summarization, 
the authors of WaveCoder evaluated MagiCoder-DS on HumanEvalExplain. According to them, 
MagiCoder-DS achieves \textbf{55.5\%} pass@1 performance on Python and \textbf{40.7\%} 
across multiple languages.

\subsection{WaveCoder}

% Summary
CodeOcean \cite{yu2024wavecoder} is a versitile dataset for fine-tuning LLMs,
containing 20,000 instruction instances across four universal code related
tasks. The authors released 3 WaveCoder models, LLMs fine-tuned on CodeOcean.
They use StarCoder-15B \cite{li2023starcoder}, CodeLLaMa-7B and 13B
\cite{rozière2023code}, and DeepSeekCoder-6.7B \cite{guo2024deepseekcoder} as
the base models of WaveCoder variants. The models are evaluated on HumanEval and
MBPP benchmarks achieving pass@1 results of up to \textbf{64.0\%} and \textbf{62.8\%}
respectively. On the HumanEvalExplain benchmark, the best WaveCoder model
reaches \textbf{48.2\%} pass@1 performance on Python and \textbf{41.3\%} across multiple
languages.

% Code Generation
The four WaveCoder models have been evaluated on two code generation benchmarks:
HumanEval and MBPP. The authors report improved performance for each model
compared to the base models. The results are in
\Cref{tab:WaveCoderHumanEvalMBPP}.

\begin{table}[ht]
    \begin{center}
        \begin{tabular}{ l | l | c | c }
            \makecell[c]{Model} & \makecell[c]{Base Model} & \makecell{HumanEval \\ (pass@1)} & \makecell{MBPP \\ (pass@1)} \\ \hline \hline
            WaveCoder-SC-15B & StarCoder & 50.5\% & 51.0\% \\ \hline
            WaveCoder-CL-7B & CodeLLaMa &  48.1\% & 47.2\% \\ \hline
            WaveCoder-CL-13B & CodeLLaMa & 55.4\% & 49.6\% \\ \hline
            WaveCoder-DS-6.7B & DeepSeekCoder & 64.0\% & 62.8\% \\
        \end{tabular}
    \end{center}
    \caption{WaveCoder models and their performance on the HumanEval and MBPP benchmarks}
    \label{tab:WaveCoderHumanEvalMBPP}
\end{table}

One of the fine-tuning objectives of the WaveCoder models was code
explanation/summarization. This aspect of the models was evaluated on
HumanEvalExplain. Most WaveCoder models outperformed most of the open-sourced
code LLMs upon publication. The results on this benchmark can be seen in
\Cref{tab:WaveCoderHumanEvalExplain}.

\begin{table}[ht]
    \begin{center}
        \begin{tabular}{ l | l | c | c }
            \makecell[c]{Model} & \makecell[c]{Base Model} & \makecell{Python \\ (pass@1)} & \makecell{Average \\ (pass@1)} \\ \hline \hline
            WaveCoder-SC-15B & StarCoder & 37.1\% & 30.8\% \\ \hline
            WaveCoder-CL-7B & CodeLlama & 41.4\% & 32.4\% \\ \hline
            WaveCoder-CL-13B & CodeLkama & 45.7\% & 37.9\% \\ \hline
            WaveCoder-DS-6.7B & DeepSeekCoder & 48.2\% & 41.3\% \\
        \end{tabular}
    \end{center}
    \caption{WaveCoder models and their performance on the HumanEvalExplain benchmark}
    \label{tab:WaveCoderHumanEvalExplain}
\end{table}

\subsection{Llama3}

A partially published family of models is Llama3 \cite{llama3modelcard}. Although these are not
fundamentally coding LLMs, they still show promising performance as they have
been trained on 4x more programming-related content compared to Llama2. So far,
two model variants have been published, with 8B and 70B parameters. Some other
variants are still under training, including one model with 400B parameters.

Based on the reported results on the HumanEval benchmark, the 8B and 70B models
achieve \textbf{62.2\%} and \textbf{81.7\%} zero-shot pass@1 performance respectively. Although
the training of the 400B variant has not yet been completed, results have been
reported based on an early checkpoint (15th of April, 2024): \textbf{84.1\%}. Results on
the HumanEval benchmark indicate that Llama3 outperforms every other opened model in
code generation. 

Although HumanEval was the only benchmark related to coding with reported results, we 
have evaluated one of the published Llama3 models on the HumanEvalExplain benchmark. 
Llama3-8B-Instruct reaches \textbf{42.7\%} pass@1 performance on Python.

\section{Conclusion}
\label{conclusion}

This technical report provides a review of the performance of some of the
leading open-sourced coding Large Language Models in text-to-code and
code-to-text tasks. As we stated earlier, our focus is code-to-text
(summarizing/explaining source code), on which less research has been done. Of
the two benchmarks reviewed, CodeXGLUE and HumanEvalExplain, HumaEvalExplain
appears to be the more widespread and acknowledged one. A summary of all the
results on HumanEvalExplain is shown in \Cref{tab:HumanEvalCompare}.

\begin{table}[ht]
    \begin{center}
        \begin{tabular}{ l | c | c }
            \makecell[c]{Model} & \makecell{HumanEvalExplain \\ (Python)} & \makecell{HumanEvalExplain \\ (Average)} \\ \hline \hline
            CodeLlama-instruct-7B & 33.5\% & 27.3\% \\ \hline
            CodeLlama-instruct-13B & 40.2\% & 28.2\% \\ \hline
            OctoGeeX-6B & 30.4\% & 22.9\% \\ \hline
            OctoCoder-16B & 35.1\% & 24.5\% \\ \hline
            WizardCoder-15B & 32.5\% & 27.5\% \\ \hline
            DeepSeekCoder-6.7B & 43.9\% & 34.6\% \\ \hline
            MagiCoder-DS-6.7B & \textbf{55.5\%} & 40.7\% \\ \hline
            WaveCoder-SC-15B & 37.1\% & 30.8\% \\ \hline
            WaveCoder-CL-7B & 41.4\% & 32.4\% \\ \hline
            WaveCoder-CL-13B & 45.7\% & 37.9\% \\ \hline
            WaveCoder-DS-6.7B & 48.2\% & \textbf{41.3\%} \\ \hline
            Llama3-8B-instruct & 42.7\% & - \\ \hline
        \end{tabular}
    \end{center}
    \caption{All reported results on HumanEvalExplain}
    \label{tab:HumanEvalCompare}
\end{table}

After comparing the code explanation performance of models on the
HumanEvalExplain benchmark, MagiCoder (DS-6.7B) demonstrated the best code
explaining capabilities in Python and WaveCoder (DS-6.7B) was the best across
multiple languages.

Unfortunately, many authors have not reported any results on source code
summarization. One reason for this could be the poor quality of the models'
summarization capabilities. This is supported by the experience of the authors
of OctoCoder, who reported 0.0 zeroshot pass@k performance on HumanEvalExplain
for two models (CodeGeeX2 \cite{zheng2023codegeex} and StarCoder
\cite{li2023starcoder}).

\newpage

\bibliographystyle{acm}
\bibliography{references}

\end{document}